\documentclass[conference]{IEEEtran}
\IEEEoverridecommandlockouts
\usepackage{cite}
\usepackage{amsmath,amssymb,amsfonts}
\usepackage{algorithmic}
\usepackage{graphicx}
\usepackage{textcomp}
\usepackage{xcolor}
\def\BibTeX{{\rm B\kern-.05em{\sc i\kern-.025em b}\kern-.08em
    T\kern-.1667em\lower.7ex\hbox{E}\kern-.125emX}}

\usepackage{booktabs}
\usepackage{dsbda-style}

\begin{document}

\author{\IEEEauthorblockN{Justin Mücke}
\IEEEauthorblockA{
\textit{University of Ulm}\\
Ulm, Germany \\
justin.muecke@uni-ulm.de}
\and
\IEEEauthorblockN{Ansgar Scherp}
\IEEEauthorblockA{
\textit{University of Ulm}\\
Ulm, Germany \\
ansgar.scherp@uni-ulm.de}
}

\title{Consistency Checking of OWL Ontologies using Graph Language Models}

\maketitle

\begin{abstract}
Semantic reasoning aims to infer new knowledge from existing knowledge, with OWL ontologies serving as a standardized framework for organizing information. A key challenge in semantic reasoning is verifying ontology consistency. However, state-of-the-art reasoners are computationally expensive, and their efficiency decreases as ontology sizes grow. While classical machine learning models have been explored for consistency checking of A-Box axioms, considering T-Boxes remains unaddressed.
Large language models (LLMs) have shown promising results for natural language inference but perform poorly on logical reasoning. The recently introduced Graph Language Model (GLM) offers a way to simultaneously process graph-structured data and text. This paper proposes GLaMoR (Graph Language Model for Reasoning), a reasoning pipeline that transforms OWL ontologies into graph-structured data and adapts the GLM architecture for consistency checking. We evaluate GLaMoR on ontologies from the NCBO BioPortal repository, converting them into triples suitable for model input. Our results show that the GLM outperforms all baseline models, achieving $95\%$ accuracy, and is $20$ times faster than classical reasoners.
The Code is accessible under: \url{https://github.com/JustinMuecke/GLaMoR}
\end{abstract}


\section{Introduction}
\label{sec:introduction}   
    With the increasing complexity of knowledge representation and reasoning systems, ontologies play a vital role in structuring domain knowledge across various fields, \eg biomedical expert knowledge. The Web Ontology Language (OWL)~\cite{owl2} is a widely adopted standard for representing ontologies, allowing for the description of concepts and the relationships between them. 
    OWL 2~\cite{OWL-2} is based on the $\mathcal{SROIQ}$~\cite{SROIQ} description logic, which supports complex reasoning while maintaining logical consistency. 
    To derive additional knowledge from these ontologies, semantic reasoners are employed to infer new facts through logical entailment.\arxivonly{~These reasoners support key tasks such as classification, query answering, and consistency checking by leveraging formal logic systems for precise and reliable inference.}
    A prominent example is HermiT~\cite{HermiT}, an efficient OWL 2-compliant reasoner that outperforms many classical reasoners in consistency checking runtime~\cite{reasoner-comparison}.
    
    Despite the advancements of reasoners like HermiT, scalability remains a significant challenge due to the high computational costs associated with reasoning over large, complex ontologies~\cite{Approximate-abox-checking}.
    To address these limitations, researchers have explored alternative methods, shifting from traditional semantic reasoning to machine learning paradigms. One attempt treats consistency checking as a binary classification problem where classifiers are trained to distinguish between consistent and inconsistent ontologies~\cite{Approximate-abox-checking}. However, this approach focuses on A-Box consistency, leaving T-Box consistencies unaddressed. 
    Another line of research employs reinforcement learning to perform question-answering (QA) by reasoning over underlying knowledge graph structures~\cite{DeepPath, AttnPath, MINERVA, MultiHopKG}. This approach is similar to A-Box reasoning, focusing on finding implicit relations. For instance, given a knowledge graph that contains the facts (Freddy Mercury, singer-of, Queen) and (Jer Bulsara, mother-of, Freddy Mercury), the model should infer that Jer Bulsara is the mother of the singer of Queen~\cite{latentlypreformmultihopreasoing}. 
    Additionally, attempts have been made to perform QA-reasoning using Large Language Models (LLMs).  
    While semantic reasoners work on well-defined logic rules, LLMs are language-driven and rely on statistical patterns in text rather than formal axioms. 
    Studies have shown that while LLMs can handle reasoning with short distances in knowledge graphs, \eg the two hops in the example above, their performance declines when tasked with finding connections over longer distances~\cite{latentlypreformmultihopreasoing}.
    The Graph Language Models (GLMs)~\cite{glm} have emerged as potential solution, integrating the text interpretation capabilities of LLMs with the attention mechanisms used in graph neural networks. 
    The GLMs modify the attention matrix of the T5 language model~\cite{T5} to support the connectivity of graph vertices.
    \arxivonly{They have demonstrated success in relation classification in knowledge graphs, suggesting they could be adapted for semantic reasoning tasks, such as ontology consistency checking when interpreted as a binary graph classification problem.}
    This work proposes a novel \textbf{G}raph \textbf{La}nugage \textbf{Mo}del for \textbf{R}easoning (GLaMoR) pipeline, which transforms ontology data and trains a reasoning model by adapting the GLM architecture to the task of ontology consistency checking. 
    In summary, our contributions are:
    \begin{itemize}
        \item An ontology consistency checking pipeline, GLaMoR, adapting the Graph Language Model~\cite{glm} for
        consistency checking as a binary graph classification task. 
        We use T5-base and T5-small as the underlying language models.

 \item A novel consistency-checking dataset of ontology modules. 
        To introduce inconsistent data points, we inject axioms based on 14 anti-patterns into the ontologies~\cite{owl-antipatterns}. 
           
        \item 
        GLaMoR outperforms all classical and modern baselines. The former include logistic regression, 
        random forests, 
        SVM, 
        Naive Bayes, and a 
        WideMLP~\cite{WideMLP}\arxivonly{~applied to mean-pooled ontology embeddings}. 
        The latter the graph foundational model PRODIGY~\cite{PRODIGY}, ModernBERT~\cite{ModernBert}, LongT5~\cite{LongT5}, and zero-shot Llama 3~\cite{Llama3}.

\item 
        We compare the total runtime (training plus inference time) of the machine learning models to the consistency-checking runtime of the semantic reasoner HermiT. 
        While HermiT took 122 hours, the slowest machine-learning model took
        11 hours for the task.
            
        \item 
        In a robustness study, one kind of inconsistency was withheld from the training data to gauge the generalization capabilities of the GLM.
        The results show that the models are robust towards simple causes of inconsistency, with the global GLMs 
        being the most robust overall.
                  
    \end{itemize}

\section{Related Work}
\label{sec:relatedwork}

\arxivonly{
    We reiterate core concepts of OWL, ontology reasoning, and modularization.
    We provide insights into ontology embeddings, graph foundational models, and machine learning reasoning.
}

     \subsection{Web Ontology Language}
        OWL is a formal language developed under the oversight of the World Wide Web Consortium (W3C). 
        It is designed to model complex relationships between objects in various categories~\cite{Making_of_OWL}, \eg we can axiomatically define the structure of a university: We can first define the classes ''Class: Professor'', ''Class: Student'', and ''Class: Course''. Together with the relations ''ObjectProperty: teaches'' and ''ObjectProperty: enrolledIn'' and individuals ''Individual: Dr.Smith, Types: Professor'', ''Individual: Alice, Types: Student'' and ''Individual: AICourse, Types: ''Course''  we can model the class as: ''Dr.Smith teaches AICourse, Alice enrolledIn AICourse''. The most recent version, OWL 2, is as expressive as the $\mathcal{SROIQ}$ description logic~\cite{SROIQ}. OWL is built upon the Resource Description Framework (RDF) and primarily uses RDF's XML syntax for its representation. It consists of key components such as classes, relationships between classes, instances of classes, properties, and restrictions~\cite{OWL-Components}. 
        Additionally, ontologies in OWL can be divided into A-Box, T-Box, and R-Box. The A-Box contains the assertion components, which deal with assertions on individuals, such as ``Object A, instance of, Class C''. The T-Box handles assertions on concepts. The RBox 
        handles assertions between relations and is often considered part of the T-Box. 
        These ontologies build the basis for semantic reasoning
    \subsection{Ontology Reasoning and Modularization}
        As the construction of ontologies by experts still leaves room for error, it is imperative to devise tools that can automatically verify their structural and logical integrity. Additionally, we want to infer new knowledge from existing ontologies. From these desires, four main reasoning problems arise: 
        (1) ontology satisfiability checking, 
        (2) concept satisfiability checking, 
        (3) concept subsumption checking, and 
        (4) instance checking. 
        Since these problems can be reduced to ontology satisfiability checking, tools known as semantic reasoners are primarily built to perform this task~\cite{paper-birte}. 
        Semantic reasoners are automated reasoning tools to ensure consistency, infer relationships, and validate data against an ontology. These problems are foundational to semantic reasoning, especially in the context of the Semantic Web, where reasoners like Pellet~\cite{Pellet} or HermiT~\cite{HermiT} are used to ensure that large-scale knowledge bases remain coherent and functional. To perform the reasoning, Pellet uses a tableaux algorithm to construct a tree model out of the ontology consisting of a graph and a labeling function. Building the graph through expansion rules identifies inconsistencies if any conflict arises during this process. HermiT improves upon this concept by introducing hyper-tableau calculus to remove non-deterministic behavior from the expansion mechanism. 
        Another Reasoner is RDF\-ox~\cite{RDFox}. It performs reasoning over RDF datasets using Datalog, a rule-based query language. Datalog rules take the form of logical implications, such as $H \xleftarrow{} B_1 \land \ldots \land B_k$, where the Head $H$ is inferred if all body conditions $B_i$ hold.\arxivonly{~RDFox applies these rules iteratively to derive new facts from an OWL ontology.}
        
        As the use of machine learning paradigms imposes size constraints on the ontologies, we need a way to reduce their total size. 
        The goal of ontology modularization is to divide a given ontology $\mathcal{O}$ into modules $\mathcal{M} = \{\mathcal{M}_1, \ldots,  \mathcal{M}_n\}$. 
        Each module should be self-contained, consistent, and topic-centric~\cite{partitioning_bioportal_ontologies}. 
        Known approaches to ontology modularization include PATO~\cite{PATO} and OAPT~\cite{OAPT}. 
        PATO works by creating a graph structure that represents dependencies between elements in the OWL ontology. This graph then gets partitioned into sets of ontology elements, which are the basis for creating modules. OAPT leverages seeding-based clustering to find root nodes of the different modules and iteratively expands these roots. 

\subsection{Embedding Models for Ontologies}
        For many machine learning tasks, it is imperative to embed the provided data into a $n$-dimensional vector space. For text and graphs, embeddings should ensure that co-occurring words or neighboring nodes have similar vector representations. 
        In OWL ontologies, it is additionally vital to maintain semantic meaning. 
        If two objects are semantically relevant to one another, their vectors should also be close. The first attempts to solve this were motivated by the embedding of knowledge graphs (KGs). 
        This includes translational models, semantic matching models, neural network models, and path-based models~\cite{embeddings}. 
        Translational models measure the plausibility of a fact as the distance between two entities. 
        In particular, TransE~\cite{TransE} represents entities and relations in the same space and aims to guarantee that if $(s,r,o)$ is part of the KG, then $h_s + h_r \sim h_o$ holds in the vector space. Semantic Matching Models rely on semantic similarity to define their scoring function. 
        This includes DistMult~\cite{DistMult}, which focuses on complex-valued embeddings, enabling it to capture symmetric and antisymmetric relations. 
        Neural networks employ the representation learning abilities of deep learning models. 
        ConvE~\cite{ConvE} uses two-dimensional convolutions to predict missing links in the KG. 
        RDF2Vec~\cite{RDF2Vec} is one of the path-based methods, which relies on random walks to generate node and edge sequences to use with a skip-gram model to compute the embeddings. 
        As RDF2Vec struggles to capture logical axioms of OWL Ontologies, OWL2Vec*~\cite{owl2vec} was developed. It explicitly incorporates the logical semantics of OWL ontologies through reasoning-based materialization, adding inferred axioms to the graph. \arxivonly{In particular, it creates an ontology graph that combines the triples in the ontology, reasoned triples obtained by using an OWL reasoner, and literal nodes linked to entities, enabling textual information to influence the embedding.}

\subsection{Graph Foundational Models}        
        Graph foundational models aim to generalize graph learning by pre-training models on diverse graph data and enabling task-specific fine-tuning~\cite{GraphFoundationalModelSurvey}. 
        One such architecture is the  Graph Language Model (GLM). 
        Language models excel at reasoning over text but struggle to handle graph-structured data.
        The GLM aims to combine the text-processing capabilities of language models with the graph-handling capabilities of graph transformers by initializing a Graph Transformer~\cite{GraphTransformer} with parameters from a T5 model~\cite{T5}, thereby integrating graph priors with the language understanding of T5~\cite{glm}. 
        To apply this architecture to knowledge graphs, it is converted into a Levi graph, a transformation that replaces each edge with a node containing the relation name and connects this node to the original head and tail nodes.\arxivonly{~This transformation helps maintain the structure of the triplets as sequences of tokens facilitated by relative positional encodings. These encodings preserve the relational structure within the transformed graph, enabling the GLM to reason over the combined graph and text data effectively.} The GLM supports two modes of operation: global GLM and local GLM, referring to the considered positional encodings. While in the global setting, the self-attention mechanism can connect any node to any other node, the local setting restricts the self-attention to tokens of the same triplet.
        Another foundational model, PRODIGY~\cite{PRODIGY}, explores a different approach by enabling in-context learning for graph tasks. 
        In-context learning enables models to generalize to new tasks by reasoning over prompts instead of updating their weights. 
        PRODIGY extends this concept to graphs by introducing prompt graphs. 
        \arxivonly{ These provide a unified way to represent diverse graph-based tasks and generalize across various tasks without fine-tuning.
        The prompt graphs consist of data graphs and task graphs.} 
        While the data graphs consist of prompt examples, the task graph connects each data point with each label with edges indicating true or false binary labels. PRODIGY employs attention-based GNNs to propagate information between data nodes and label nodes.\arxivonly{~It reasons over relationships in graph data and effectively generalizes across tasks without fine-tuning.} 
        PRODIGY is used as a baseline from the family of graph foundation models, which also includes models like PSP~\cite{PSP} and GraphAny~\cite{GraphAny}.

\subsection{Machine Learning Reasoning}

        Prior work investigated classical machine learning models for A-Box consistency checking in ontologies, including Decision Trees, Naive Bayes, SVMs, and Random Forests~\cite{Approximate-abox-checking}. While the potential of machine learning paradigms for consistency checking was demonstrated, they focused solely on the A-Box, ignoring the T-Box, \ie only relation assertions and all the type assertions for the subject and object were considered, not considering the full semantics of the underlying ontologies.
        Another approach employed reinforcement learning for knowledge base completion. Methods such as DeeptPath~\cite{DeepPath}, AttnPath~\cite{AttnPath}, MINERVA~\cite{MINERVA}, and MultiHopKG~\cite{MultiHopKG} train an agent to explore reasoning paths in a knowledge graph. The main goal is to optimize a reward function to retrieve multi-hop relationships between entities. While this approach is effective in inferring implicit facts, they are not suitable for consistency checking. The main drawback of these approaches lies in the locality of the task. In question-answering, the required information can be found over short relational paths, \eg 1-hop or 2-hop reasoning, while consistency checking requires a global view of all relations and restrictions, ensuring that no contradiction exists between different axioms. This is particularly crucial for detecting cardinality violations or existential quantification, as these extend beyond simple entity relationships.
        Similarly, LLMs have been explored for multi-hop reasoning tasks in natural language. 
        LLMs reach reasonable accuracies for 1-hop reasoning but struggle with multi-hop reasoning as complexity increases. Their main limitation is the reliance on statistical associations rather than logical inference~\cite{latentlypreformmultihopreasoing}. 
        Thus, LLM-based reasoning lacks an explicit representation of logical constraints. 
        As reinforcement learning approaches, the investigated capabilities of the LLMs are bound to a local area around subjects in question-answering. 
        Thus, they cannot discern violations of restrictions multiple hops apart in the underlying knowledge base.

\begin{figure*}[th!]
        \centering
        \includegraphics[width=0.9\textwidth]{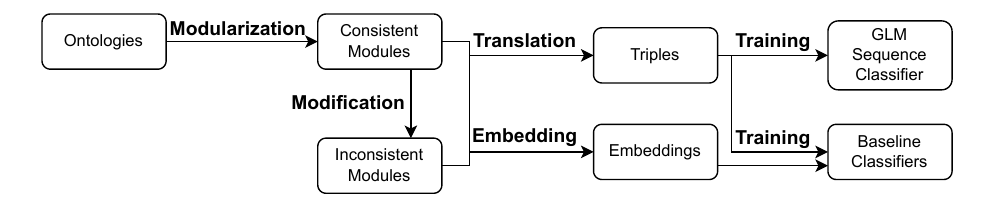}
        \caption{GLaMoR pipeline of acquiring and preprocessing the data and using it to train the graph language model (GLM).}
    \label{fig:pipeline}
    \end{figure*}

\section{GLaMoR Pipeline}
\label{sec:methodology}
    
We adopt the Graph Language Model architecture~\cite{glm} to classify whether OWL ontologies are consistent. This section explains the steps to transform our dataset into the necessary format. 
\arxivonly{The data we use is provided by the BioPortal repository, which provides access to all ontologies via a REST API.}
The steps to process the data can be seen in \autoref{fig:pipeline}.

    \subsection{Modularization}
    \label{modularization}
        BioPortal~\cite{BioPortal}, an online ontology library, contains $1,143$ ontologies, which hold a total of $15,598,525$ classes. With the ontologies containing an average of $13,647$ classes alone, the need for some form of modularization arises, as the maximal number of triples that can be processed is limited by the size of the GLM input. To this end, we use the Ontology Analysis and Partition tool (OAPT)~\cite{OAPT}, which follows four key steps to generate modules: (1) ranking ontology concepts, (2)~determining cluster heads, (3) partitioning the ontology, and (4) generating the Module. To rank the ontology concepts, the importance of each concept is quantified within the ontology. The resulting important concepts can then be used as cluster heads, which form the basis of the partition. When determining cluster heads, it is important to define the number of heads needed and which concepts should be considered. To partition the ontology, one partition is initiated for each cluster head. Each direct child of a head is then placed in the corresponding cluster, and the remaining concepts that have thus far not been assigned to a cluster get sorted into fitting partitions using a membership function. The resulting clusters of concepts are then used to create the modules while ensuring the intra-relationship between concepts in the same partition.

    \subsection{Modification}
    \label{modification}
        To train our model to distinguish between consistent and inconsistent OWL ontologies, we require a diverse set of inconsistent examples. To our knowledge, no dedicated repository exists, so we generate inconsistencies by injecting axioms into existing ontologies. To avoid bias towards a single inconsistency type, we introduce a well-balanced dataset covering various inconsistency sources.
        We systematically create inconsistencies using OWL anti-patterns, which define axioms that lead to logical contradictions. 
        A pattern usually refers to a guideline on how to properly form ontologies. Thus, anti-patterns describe ways to systematically create faulty ontologies. Specifically, we base our approach on previously defined logical patterns~\cite{owl-antipatterns} and introduce three additional ones. 
        The Out of Range and Out of Domain anti-patterns occur when an individual is assigned to a property in a way that violates the ontology’s class constraints. Specifically, the individual is an instance of a class that is disjoint with the expected domain or range of the property, leading to inconsistency. The cyclic subclass anti-pattern describes a cyclic structure in the subclasses. Given an anti-pattern of $n$ axioms, we verify whether the ontology we want to introduce inconsistencies into contains $n-1$ or $n-2$ of these axioms. If they do, we inject the missing axioms to complete the anti-pattern. Thus, we can create inconsistent ontologies programmatically. 
        The anti-patterns used are listed in \autoref{tab:anti-pattern}.

        \begin{table}[ht!]
            \centering
            \caption{List of anti-patterns to form logical inconsistencies. While the first ones have been previously defined~\cite{owl-antipatterns}, the lower three are introduced by us.}
            \vspace{3mm}
            \begin{tabular}{ll}
            \toprule
                Name                                    &  Pattern \\\midrule
                AndIsOr (AIO)                           & $c_1 \sqsubseteq \exists R.(c_2 \sqcap c_3)$,                         \\
                                                        & $\operatorname{Disj}(c_2,c_3)$                                        \\
                EquivalenceIsDifference (EID)           & $c_1 \equiv      c_2$,                $\operatorname{Disj}(c_1, c_2)$ \\
                OnlynessIsLoneliness (OIL)              & $c_1 \sqsubseteq \forall R.c_2$,      $c_1\sqsubseteq\forall R.c_3$,  \\
                                                        & $\operatorname{Disj}(c_2, c_3)$                                       \\
                OILWithInheritence (OILWI)              & $c_1\sqsubseteq\forall R.c_3$,        $c_2\sqsubseteq\forall R.c_4$, \\
                                                        & $c_1 \sqsubseteq c_2$,                $\operatorname{Disj}(c_3,c_4)$  \\
                OILWithPropertyInheritance              & $c_1\sqsubseteq\forall R2.c_3$,       $c_1\sqsubseteq\forall R_1.c_2$,\\
                (OILWPI)                                & $R_1 \sqsubseteq R_2$,                $\operatorname{Disj}(c_2,c_3)$  \\
                UniversalExistance(UE)                  & $c_1 \sqsubseteq \forall R.c_2$,      $c_1\sqsubseteq\exists R.c_3$,  \\
                                                        & $\operatorname{Disj}(c_2, c_3)$                                       \\
                UEWithInheritance1                      & $c_3 \sqsubseteq \forall R.c_4$,      $c_1\sqsubseteq\exists R.c_3$,  \\
                (UEWI\_1)                               & $c_1 \sqsubseteq c_2$,                $\operatorname{Disj}(c_3, c_4)$ \\
                UEWithInheritance2                      & $c_2 \sqsubseteq\exists R.c_4$,       $c_1\sqsubseteq\forall R.c_3$,  \\
                (UEWI\_2)                               & $c_1 \sqsubseteq c_2$,                $\operatorname{Disj}(c_3, c_4)$ \\
                UEWithPropertyInheritance               & $c_1 \sqsubseteq\forall R_2.c_3$,     $c_1\sqsubseteq\exists R_1.c_2$,\\
                (UEWPI)                                 & $R_1 \sqsubseteq R_2$,                $\operatorname{Disj}(c_2,c_3)$  \\
                UEWithInverseProperty                   & $c_2 \sqsubseteq \exists R^{-1}.c_1$, $c_1\sqsubseteq\forall R.c_3$,  \\
                (UEWIP)                                 & $\operatorname{Disj}(c_2, c_3)$                                       \\
                SumOfSomeIsNeverEqualToOne              & $c_1 \sqsubseteq \exists R.c_2$,      $c_1\sqsubseteq\exists R.c_3$,  \\
                (SOSINETO)                              &  $c_1\sqsubseteq \leq 1.T$,           $\operatorname{Disj}(c_2, c_3)$ \\
                \midrule
                OutOfDomain (OOD)                       & $c_1 = \operatorname{Domain}(p)$,                    $p(a,b)$,                       \\
                                                        &  $a \in c_2$,                         $\operatorname{Disj}(c_1, c_2)$ \\
                OutOfRange (OOR)                        & $c_1 = \operatorname{Range}(p)$,                     $p(a,b)$,                       \\
                                                        & $b \in c_2$,                          $\operatorname{Disj}(c_1,c_2)$  \\
                CyclicSubClass (CSC)                    & $c_1 \sqsubseteq c_2$,                $c_2 \sqsubseteq c_3$,          $c_3 \sqsubseteq c_1$ \\
                \bottomrule
                
            \end{tabular}
            \label{tab:anti-pattern}
        \end{table}

\subsection{Translation}
        The modules generated by the OAPT are saved using the XML/RDF syntax. As this syntax contains all prefixes in place and its main goal is not to be human-readable, we translate the ontologies into the Manchester syntax. This syntax has been primarily developed to be human-readable by moving the prefixes of the classes and relations to the top of the ontology, thus providing a simpler and more structured format. We use the data in Manchester syntax to further transform the ontologies into a set of triples $(subject, relation, object)$. 
    These can be used directly as input for our GLM. 
    To exploit the qualities of the language model, the triples need to form correct sentences. Thus, we first remove the prefixes that identify the source from which the object originates, \eg RDFs or OWL. In addition, we translate the structures into English.
    \shortorextended{Examples can be found in our technical report~\cite{DBLP:journals/corr/abs-2504-19023-glamor}.}{Examples can be found in Appendix~\ref{app:Translation}.} 

    \subsection{Embedding}
        As models such as logistic regression and PRODIGY require embeddings of the ontologies as input, we embed our modules using OWL2Vec*~\cite{owl2vec}. 
        To embed an OWL ontology, OWL2Vec* follows several steps: (1) Extraction of an RDF Graph, (2) Random Walks over the Graph, (3) Extraction of lexical Information, (4) Creation of a structure and lexical Document, (5) Combining the documents, and (6) Training a word2vec model~\cite{word2vec} using the combined document as input. 
        To extract the RDF graph, they transform simple axioms according to the mapping provided by the W3C, and the complex ones are split into multiple triples. 
        The lexical information contains information about the labels and comments in the ontology. 
        By performing random walks over the RDF graph, they create sentences consisting of entity URIs, whereas the lexical document contains sentences of words. 
        The combination of these can then be used to apply NLP methods for embedding. 
        
    \subsection{Training}
        Using a sequence of triples for each module, along with a label indicating whether the sequence is consistent or inconsistent, we train a GLM to perform a binary classification. Thus, we need to adapt the proposed GLM architecture, as it was originally designed to perform masked token prediction~\cite{glm}. We achieve this by adapting the model's output not to predict the logits of a singular input token but to output the average of all logits. Additionally, we apply mean pooling to the embeddings and use them as input for a logistic regression, which serves as a baseline for our experiments. We also use the embeddings created by OWL2Vec* to train a PRODIGY model and a WideMLP to compare to the performance of GLM. Lastly, we use the triples as a continuous text to fine-tune a LongT5 and a ModernBERT model. Further, we use a Llama 3 model in a zero-shot setting. 
        \shortorextended{The prompt for Llama 3 can be found in the technical report~\cite{DBLP:journals/corr/abs-2504-19023-glamor}.}{
        The prompt for Llama~3 can be found in Appendix~\ref{sec:prompt}.       
        }

\section{Experimental Apparatus}
\label{sec:experimentalapparatus}
    \arxivonly{
    We describe the creation of our dataset, preprocessing statistics, hyperparameter optimization, the procedure, and the metrics.
    }
    \subsection{Dataset}
    \label{sec:datasets}
       \arxivonly{As this paper aims to create models to perform binary classification on OWL ontologies based on their consistency, 
       We require a dataset that consists of OWL ontologies and contains consistent and inconsistent data points. 
        Additionally, we want to include as many different kinds of relationships and structures as possible. 
        We decided not to include datasets like DBpedia~\cite{dbpedia}, which has been used for A-Box reasoning~\cite{Approximate-abox-checking}, as it primarily consists of knowledge graph structures with minimal ontological constraints.}
        Since no existing dataset provides both structured OWL ontologies and known inconsistencies, we constructed our own. 
        As basis for the dataset, we use the ontologies provided by the NCBO BioPortal repository~\cite{BioPortal}. We could access $1,054$ of the provided ontologies with $15,604,437$ classes and $36,286$ properties. Further statistics on these ontologies can be found in Table~\ref{tab:bioportal}. As the GLM limits the maximum sequence length that can be used as input, we need to decrease the size of these ontologies. Applying the modularization described in Section~\ref{modularization}, we could translate the ontologies into $7,505$ consistent modules. On average, a single ontology could be split into $7.74$ modules. 
        \arxivonly{More detailed information is available in Appendix~\ref{app:Modularization}.}
        In Table~\ref{tab:bioportal}, we can also see that the median number of classes decreased by $29.82\%$ and the median number of properties decreased by $83.33\%$ due to the modularization. Furthermore, using the modification rules defined in Section~\ref{modification}, we created $19,902$ inconsistent modules. 
        \shortorextended{Details on which anti-pattern was injected can be found in our technical report~\cite{DBLP:journals/corr/abs-2504-19023-glamor}.}{Details on which anti-pattern was injected can be found in Appendix~\ref{app:Modification}.}
        
        \begin{table}[h]
            \centering
            \caption{Number of OWL classes and properties in the ontologies of the NCBO BioPortal and their breakdown in modules.}
            \vspace{3mm}
            \begin{tabular}{l|rrr}
            \toprule
                \textbf{$1,054$ Ontologies}     & Median      & Average     & SD     \\\hline
                OWL Classes                     & $295.00$    & $10,039.99$ & $53,620.56$           \\
                Properties                      & $12.00$     & $119.07$    & $2132.41$            \\
            \midrule
                \textbf{$7,505$ Modules}        & Median      & Average     & SD    \\\hline
                OWL Classes                     & $210.00$     & $1,770.57$ & $11,391.92$              \\
                Properties                      & $2.00$       & $27.03$    & $621.40$\\ 
            \bottomrule
            \end{tabular}
            
            \label{tab:bioportal}
        \end{table}

\arxivonly{\subsection{Preprocessing}}
    \label{sec:preprocessing}
        We perform some preprocessing before submitting our data points into the pipeline. The downloaded ontologies contain axioms whose task is to improve human readability. As we are only interested in the semantics of the underlying ontologies, we exclude these axioms. This includes annotation axioms like SKOS and OBO axioms, which are mainly used to link knowledge between ontologies, as well as description and label axioms, which aim to improve readability. 

\subsection{Hyperparameter Optimization}
    \label{sec:hyperparameter}
        First, we created the embeddings using the OWL2Vec* algorithm with an embedding size of $100$, $10$ iterations to train the underlying Word2Vec model, and a window of $5$.
        For optimizing the decision tree, logistic regression, random forest, SVM, and naive Bayes, we perform a grid search over the respective parameters. 
        For each model, we use $85\%$ of the data to perform a 10-fold cross-validation and pick the best model based on its accuracy on the remaining $15\%$ of the data.
        The decision tree performed best with a maximum depth of $5$, minimum samples per leaf of $5$, and a minimum samples split of $20$, with entropy as the splitting criterion. For the logistic regression, we arrived at a regularization strength of $0.03359$, L2-Regularization as a penalty, and a maximum number of iterations of $50$ with the L-BFGS optimization algorithm~\cite{L-BFGS}.
        The random forest performed best with a maximum depth of 9, no maximum features, and 20 or 100 estimators with a maximum of 20 or 100 leaf nodes. Additionally, SVM performed best with the RBF kernel and a cost of 1. Lastly, the Naive Bayes used a smoothing of $10^{-9}$. 
        We applied a $(70,15,15)$ data split along the ontology modules.
        Hyperparameter optimization via grid search resulted in a learning rate of $1\cdot 10^{-4}$, weight decay of $5\cdot10^{-4}$, and the weight parameter set to $45$.
        Using the WideMLP, we achieved the best performance when the learning rate is set to $5\cdot 10 ^{-5}$, the weight decay set to $0$, and the dropout to $0.5$.
        For ModernBERT, the grid search resulted in a learning rate of $1\cdot 10^{-5}$ and a weight decay of $1\cdot10^{-4}$.
        The LongT5 had the best performance with a learning rate of $5\cdot 10 ^{-5}$ and a weight decay of 0.
        For the GLMs, guided by the original paper, we tested different learning rates with a maximum of $50$ epochs and early stopping. 
        All GLMs performed best with a learning rate of $10^{-4}$.

\subsection{Procedure}
    \label{sec:procedure}
        
        First, we create a balanced dataset out of the total amount of data points we acquired. As we want to train the GLM models, we additionally have to adhere to the maximum length of the input sequence permitted by the model. As the T5 model is restricted to a sequence length of at most $4,096$ tokens, this is the threshold we use to filter the dataset. This filter is needed as the model must process the whole ontology to classify it. With the remaining data points, we balance the consistent and inconsistent examples. For the anti-patterns EID, CSC, UE, and AIO, we use a randomly sampled fraction of them, where the original distribution among them is upheld, as we have a greater number of data points of these patterns as compared to the rest. The resulting distribution of examples can be seen in \autoref{tab:dataset}. We then create a $(70,15,15)$ train/validation/test split. To train the classic machine learning models, we require graph embeddings of the modules; thus, we embed the modules using the OWL2Vec* algorithm and mean-pool the feature vectors. Additionally, these embeddings are used to train a WideMLP as well as PRODIGY, which uses the Adam optimizer. We translate the modules into triples to train the GLM using the AdamW optimizer. We use combinations of the local and global settings paired with the T5-small and T5-base models as the underlying models. Furthermore, we combine the triples into a single text with proper punctuation to fine-tune a LongT5 as well as a ModernBERT model. We also use the text to test Llama 3 in a zero-shot setting. 
        We use the classical reasoner HermiT to perform consistency checking on the test set to compare its runtime to the models. 
        
        Lastly, we perform a robustness study investigating how well the graph language models handle missing data during training. Thus, we group our inconsistencies by their underlying base error, \eg OIL, OILWI, and OILWPI are considered as the group OIL*. For each group, we train the models again with those excluded from the training data and analyze how the accuracy, precision, and recall shift on the test data.
        \begin{table}[ht]
            \centering
            \caption{Modules filtered by token length ($\le 4,096$ tokens). 'Total' reflects the number of modules adhering to the size restriction, and 'Used' is the number of modules that are kept.}
            \vspace{3mm}
            \begin{tabular}{l|r|r}
                \toprule
                    Number of modules       & Total     & Used \\
                \toprule
                    Consistent modules      & $4,169$   & $4,169$ \\
                \midrule
                    Inconsistent modules    & $10,079$  & $4,169$ \\
                \hline
                    AIO                     & $1,338$   & 538 \\
                    EID                     & $3,656$   & $1,467$ \\
                    OIL                     & 1         & 1 \\
                    OILWI                   & 1         & 1 \\
                    OILWPI                  & 0         & 0 \\
                    UE                      & $1,357$   & 544 \\
                    UEWI1                   & 63        & 63 \\
                    UEWI2                   & 62        & 62 \\
                    UEWPI                   & 4         & 4 \\
                    UEWIP                   & 20        & 20 \\
                    SOSINETO                & 6         & 6 \\
                    OOR                     & 27        & 27 \\
                    OOD                     & 24        & 24 \\
                    CSC                     & $3,520$   & $1,412$ \\
                \bottomrule
            \end{tabular}
            \label{tab:dataset}
        \end{table}
    \subsection{Metrics}
    \label{sec:measures}
        We use binary accuracy, recall, and precision for sequence classification. Accuracy measures the overall proportion of correct predictions the model makes, including consistent and inconsistent ontologies. It provides a general indication of how well the model correctly classifies both classes. Recall evaluates how effectively the model identifies inconsistent ontologies. It calculates the proportion of actual inconsistencies that the model correctly classifies as inconsistent. High recall means the model is good at detecting inconsistencies. Precision measures the proportion of predicted inconsistent ontologies that are inconsistent. It reflects how accurate the model is when it classifies an ontology as inconsistent. A high precision corresponds to the model making fewer false positive errors.
        
\section{Results}
\label{sec:results}

\begin{table*}[!ht]
        \centering
        \caption{Model performance. 
        For GLMs, 'Small/Base' indicates the T5 model size. The best result by a machine learning model per metric is bold, the second best is cursive, and the third best is underscored. For the metrics, the average of five runs is reported. The standard deviation is reported in the subscript. For the training and inference time, only the average is reported.\newline        
        Note that inference is performed on the ontology modules, both by the machine learning models as well as the classical reasoner HermiT.
        Thus,preprocessing and ontology modularization do not affect the results.
        }
            \vspace{3mm}
        \begin{tabular}{lccccr}
        \toprule
        Model           & Accuracy          & Precision         & Recall             & Training      & Inference \\\midrule
        Decision Tree   & $56.11_{(0.99)}$  & $56.94_{(00.79)}$ & $48.00_{(08.56)}$  & 00:00:00.44    & 0:00:00.00    \\
        LR              & $60.64_{(0.99)}$  & $61.89_{(01.89)}$ & $57.48_{(01.66)}$  & 00:00:00.02    & 0:00:00.00    \\
        Random Forest   & $60.50_{(2.23)}$  & $61.71_{(02.49)}$ & $56.97_{(02.90)}$  & 00:00:26.34    & 0:00:00.01    \\ 
        SVM             & $59.70_{(1.18)}$  & $61.47_{(01.52)}$ & $54.15_{(01.52)}$  & 00:00:01.56    & 0:00:00.47    \\
        Naive Bayes     & $60.64_{(1.39)}$  & $61.98_{(02.33)}$ & $54.13_{(02.18)}$  & 00:00:00.00    & 0:00:00.00     \\ 

        \midrule
        
        PRODIGY         & $50.00_{(0.38)}$  & $62.19_{(35.89)}$ & $\textbf{98.47}_{(01.95)}$  & 00:06:47.38    &  0:00:13.44   \\ 
        WideMLP         & $61.18_{(1.68)}$  & $61.47_{(01.33)}$ & $61.18_{(01.68)}$  & 00:00:14.64    &  0:00:00.07   \\
        ModernBERT      & $\underline{94.27}_{(2.09)}$  & $\textit{95.82}_{(02.31)}$ & $92.65_{(02.40)}$  & 04:57:40.57    &  0:04:07.12   \\
        LongT5          & $75.75_{(4.99)}$  & $79.36_{(08.18)}$ & $72.24_{(15.17)}$  & 11:07:55.30   &  0:04:41.19   \\
        Llama 3 0-shot   & $52.53_{(0.20)}$  & $55.48_{(00.11)}$ & $40.04_{(00.23)}$  &    ---        & 1:13:23.00    \\
        
        \midrule
        
        $\ell$GLM-small & $86.47_{(0.97)}$  & $86.55_{(02.13)}$ & $86.67_{(04.13)}$  & 04:29:11.36    &  0:14:32.12   \\
        $\ell$GLM-base  & $91.26_{(0.97)}$  & $93.27_{(02.18)}$ & $89.46_{(01.12)}$  & 04:27:56.27    &  0:16:53.06   \\
        gGLM-small      & $\textit{94.80}_{(0.43)}$  & $\underline{95.28}_{(01.18)}$ & $\textit{94.68}_{(01.07)}$  & 05:45:48.10    &  0:15:30.32   \\
        gGLM-base       & $\textbf{95.13}_{(1.10)}$  & $\textbf{96.10}_{(01.97)}$ & $\underline{94.17}_{(01.87)}$  & 04:28:15.21    &  0:17:51.30   \\
        \toprule
        \midrule
        HermiT          & 100               & 100               & 100               &     ---        & 122:24:32.42  \\
        \bottomrule
        \end{tabular} 
        \label{tab:result}
    \end{table*}

\begin{table*}
            \centering
            \caption{Model performance when a specific family of inconsistencies is excluded from the training data. A family is indicated by a "*" and represents all anti-patterns with the same base inconsistency.}
            \vspace{3mm}
            \begin{tabular}{l|ccc|ccc|ccc|ccc}
            \toprule
                           & \multicolumn{3}{c}{$\ell$GLM-small}  & \multicolumn{3}{c}{$\ell$GLM-base}  & \multicolumn{3}{c}{gGLM-small}  &\multicolumn{3}{c}{gGLM-base} \\
               Pattern     & Acc     & Pre     & Rec     & Acc     & Pre     & Rec     & Acc     & Pre     & Rec     & Acc     & Pre     & Rec \\\midrule
                AIO        & $85.21$ & $91.14$ & $78.16$ & $92.44$ & $95.58$ & $89.08$ & $95.31$ & $95.84$ & $94.77$ & $95.62$ & $96.45$ & $94.77$ \\
                EID        & $83.30$ & $90.89$ & $74.20$ & $91.33$ & $95.79$ & $86.55$ & $95.54$ & $97.36$ & $93.67$ & $95.62$ & $98.00$ & $93.19$ \\
                OIL*       & $87.36$ & $91.13$ & $82.91$ & $91.65$ & $96.30$ & $86.70$ & $95.31$ & $96.43$ & $94.14$ & $96.02$ & $97.70$ & $94.30$ \\
                UE*        & $86.24$ & $89.09$ & $82.75$ & $89.66$ & $87.18$ & $81.80$ & $90.30$ & $96.36$ & $83.86$ & $90.93$ & $97.43$ & $84.17$ \\
                SOSINETO   & $84.81$ & $88.34$ & $80.37$ & $89.74$ & $90.49$ & $88.92$ & $95.31$ & $95.40$ & $95.25$ & $96.26$ & $96.80$ & $95.72$ \\
                OO*        & $85.05$ & $82.64$ & $88.92$ & $92.36$ & $92.27$ & $92.56$ & $94.83$ & $93.95$ & $95.88$ & $95.23$ & $97.66$ & $92.72$ \\
                CSC        & $80.12$ & $92.82$ & $65.50$ & $89.82$ & $93.15$ & $86.07$ & $94.83$ & $97.64$ & $91.93$ & $95.54$ & $95.71$ & $95.41$ \\
                \bottomrule
                
                \end{tabular}
            
            \label{tab:robustness}
        \end{table*}
    \autoref{tab:result} reports the accuracy, precision, and recall of the considered models. We can see that the classical machine-learning approaches performed similarly. Out of them, 
     Logistic Regression and Naive Bayes achieved the best accuracy ($60.64$), while Logistic Regression demonstrated the best precision ($61.89$) and recall ($57.48$).
    While the PRODIGY model outperformed all models in recall, it had the worst accuracy overall. Another model that performed similarly poorly to PRODIGY was Llama 3 in the zero-shot setting, with its accuracy and precision being slightly above $50\%$, and the recall being only $40\%$, which is the worst recall of all models. Although WideMLP outperformed classical models by a few percentage points, it still lagged behind both LongT5 ($75\%$ accuracy) and ModernBERT, the strongest baseline. 
    ModernBERT showed performance similar to the global GLM with T5-small as the underlying model, reaching an accuracy of $94.27$. 
    While local GLMs outperformed classical models, they consistently underperformed compared to the global GLMs. The highest accuracy and recall were reported by the global GLM with T5-base as the underlying LLM.
    
    In terms of runtime, as shown in the last column of \autoref{tab:result}, the runtime of ML models is much lower than the runtime of classical reasoners. While the classical reasoner HermiT took over 122 hours to check the modules in the test set, the machine learning model that took the longest time to train was LongT5, with a training time of 11 hours. 
    The other modern models took less than 6 hours to train, and the classical models even below a minute.
    
    Regarding our robustness analysis, we can see in \autoref{tab:robustness} that the impact of missing data is higher for patterns with higher complexity. In particular, leaving out patterns of the UE family resulted in the highest decrease in accuracy. 
    Accuracy was barely affected with simpler patterns like AIO and CSC. 

\section{Discussion}
\label{sec:discussion}
\arxivonly{
    We discuss the key insights gained from the results, the results of the robustness study, threats to the validity of this work, and future work.
    }
    
    \subsection{Key Insights}
    \label{sec:keyresults}
        Our results indicate that incorporating global attention mechanisms strongly improves classification performance in Graph Language Models. Additionally, an increased size of the underlying model does not necessarily improve the consistency-checking capabilities. Furthermore, mean-pooled embeddings are likely unfit for the task, as they fail to preserve semantic relations between entities. The outcome of our research reveals the potential of Graph Language Models when applied to the problem of consistency checking. As can be seen in \autoref{tab:result}, the GLM in the global setting with a T5-base as the underlying model achieved the best accuracy of $95.13\%$ and the highest precision of all the models considered. Thus, it misclassified consistent ontologies as inconsistent the least, with only $3\%$ of classifications of inconsistent ontologies that are actually consistent ontologies. Meanwhile, the models using the local setting performed worse, even though the accuracy improved with the increased size of the underlying T5 model. Still, compared to the LongT5, which shares the same base model as the T5-base used in our GLM, the local GLM still performed better by $10\%$ accuracy, showing that the Graph Attention used in the GLM has a great impact on the overall consistency checking capabilities of the models.
        Another high-performing model was the fine-tuned ModernBERT which was only outperformed by the global GLMs. It showed strong results in finding the anti-patterns and achieved an accuracy of $94.27\%$. Furthermore, the classical machine learning models as well as PRODIGY and WideMLP, which rely on embeddings of the ontologies, performed noticeably worse. One factor contributing to the bad performance could be that the embedding itself did not capture the structure well. Most likely, the mean-pooling of the embeddings led to a loss of the semantic information contained in the embeddings, thus making it hard to detect inconsistencies.
        Despite near-perfect recall, PRODIGY's $50\%$ accuracy indicates it overpredicts inconsistencies. In a balanced dataset, half the data was misclassified, showing that most consistent instances were labeled as inconsistent.
        
        Furthermore, we looked into the classifications of the single patterns and found that no model struggled with any anti-pattern in particular. 
        If a model performed better than another, its performance difference was noticeable across all anti-patterns in the test data. 
        
        As seen in \autoref{tab:result}, the GLMs took the longest to train and classify the test set, with the local GLM with the T5 base taking six hours to tokenize the data and train the model. Comparing it to HermiT, which is one of the better-optimized classical reasoners~\cite{reasoner-comparison}, it took 122 hours to check the consistency of the test set. This is still a strong improvement considering we limited ourselves to smaller ontology modules. 
        \arxivonly{Global GLMs offered the best performance-to-runtime ratio, requiring under six hours to train while outperforming all other models, including ModernBERT and T5.}

\subsection{Robustness Study}
    \label{sec:robustness}
The ability of the models to handle missing data in the training data reported in \autoref{tab:robustness} shows that the global models are overall more robust than their local counterparts. 
One also sees that removing the family of UE patterns has the most impact on the performance of all the models. A reason might be that they are the more complex patterns in the data. 
Also, they do not overlap much with other patterns, as they are the only ones that contain universal ($\forall$) and existential ($\exists$) restrictions. 
The CSC pattern influenced the performance of the local models considerably. 
Patterns of this form are hard for the models in the local setting, as they are based on cycles that stretch farther than their attention, \eg CSC contains a cycle over three axioms, while the local attention is only over a single axiom. 
These problems are not noticeable in the performance of the global models, which suggests they can capture long-range dependencies. 
Overall, our results suggest that the models exhibit strong generalization capabilities, particularly in the global setting. 
The global GLM with a T5-base architecture demonstrates the highest robustness, maintaining over $90\%$ accuracy even when specific inconsistency families are withheld from training. 
This suggests that the model is able to learn broader representations that support generalization to inconsistency types not encountered during training.
\arxivonly{In other words, the global models not only outperform in raw accuracy but also generalize better when key inconsistency patterns are missing.}

    \subsection{Limitations and Future Work}
    \label{sec:threattovalidity}

        \arxivonly{
        A key challenge for our work was the unavailability and, thus, the need for the programmatic creation of inconsistent ontologies. 
        To fully capture the capabilities of graph language models, a dataset containing inconsistencies utilizing all capabilities of OWL 2 would provide better insight. Nevertheless, the inconsistencies used in this study were based on previously identified patterns that occurred repeatedly in real-world applications~\cite{owl-antipatterns}. 
        Thus, this study still provides a solid investigation into the consistency-checking capabilities of modern machine learning models. 
        Another problem might be the selection of the ontologies, as we sourced them from one repository aligned with biomedical research. While this could introduce some biases into the dataset, they should not influence the performance of the models, as all references to potential IRIs that would link them to a specific field have been removed from the data, either by the translation into triples or by the embedding of the ontologies. 
      }

    \label{sec:futurework}
        While our work already demonstrated the potential of Graph Language Models for consistency checking in OWL ontologies, further research is required to address key limitations. One of the main challenges in this study was the availability of inconsistent ontologies and the sequence length constraints imposed by GLMs. To tackle the first issue, it is imperative to establish a repository of real-world inconsistent ontologies as they emerge in practical modeling scenarios. Such a dataset would provide two main advantages: (1) it would allow for a more precise assessment of the impact on real-world applications, and (2) it would capture a wider variety of OWL 2 inconsistencies beyond the structured patterns injected programmatically in this study. 
        
        Additionally, it is worth exploring alternative large language models (LLMs) within the GLM architecture. While T5 demonstrated exceptional performance, its primary limitation remains the default input size of 512 tokens, which had to be manually increased fourfold for this study. A model with a larger context window or indefinite input length would be more scalable, allowing this approach to be applied in cases where traditional reasoners struggle with computational costs. \arxivonly{Future research should investigate whether increasing sequence length beyond our modified limit leads to continued performance improvements or whether alternative architectures (\eg retrieval-augmented or sparse attention models) could better handle long sequences in ontological reasoning.}

\section{Conclusion}
\label{sec:conclusion}   
\arxivonly{We investigated the capabilities of graph language models (GLMs) for consistency checking in OWL ontologies, which is a critical task for ensuring the logical integrity of semantic data.} \arxivonly{We developed a pipeline that transformed ontologies into smaller modules, which were then converted into triples and embedded into a vector space. Using these data points, we trained several models to evaluate their performance in consistency checking.} Our findings reveal that GLMs, particularly those trained in the global setting with graph-wide positional encoding, outperformed local counterparts as well as embedding-based models like PRODIGY, logistic regression, random forest, decision trees, SVM, and Naive Bayes. This advantage is attributed to the global GLM’s ability to capture dependencies across the entire graph, making it more effective in handling complex consistency patterns. Additionally, the global GLM is most robust to 
inconsistency types not encountered during training.
\arxivonly{These results not only show that consistency checking is no longer the exclusive domain of semantic reasoners but also suggest that machine learning models show great promise in offering scalable, high-accuracy alternatives for ontology management.}

\bibliographystyle{ieeetr}
\bibliography{references-shortened}

\clearpage

\appendix

\subsection{Modularization}
\label{app:Modularization}

\begin{figure}[h]
        \centering
        \includegraphics[width=\linewidth]{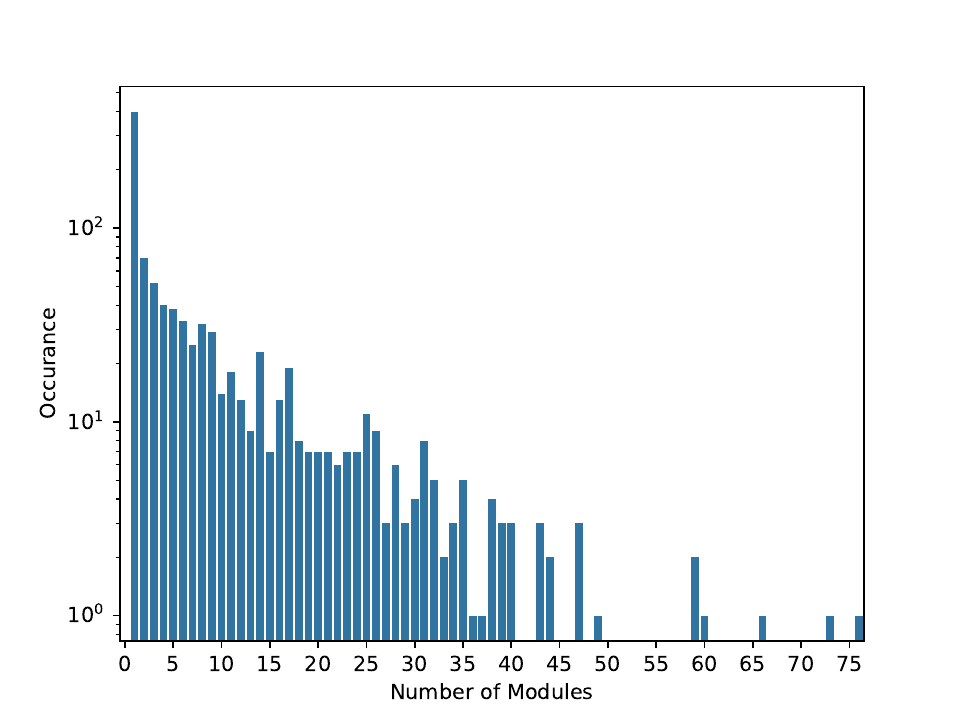}
        \caption{Distribution on the number of ontologies that can be separated into a certain number of modules.}
        \label{fig:module_distribution}
    \end{figure}

    Given the $1,054$ ontologies retrieved from the NCBO BioPortal Repository, we modularized them into $7,505$ modules. How many modules could be extracted from each ontology can be seen in \autoref{fig:module_distribution}.
    
\subsection{Modification}
\label{app:Modification}
    To create inconsistent modules, we injected axioms into the consistent modules. For each module, we tried to complete each anti-pattern. If the pattern could not be completed with one injection, we tried to complete the pattern with two injections. How often each pattern could be completed can be seen in \autoref{tab:injection}.
    
   \begin{table}[h]
        \centering
        \caption{Number of \textit{inconsistent} modules created by injecting axioms into the $7,505$ consistent modules. One Injection refers to the pattern being injected by one axiom. Two Injections refers to two axioms being injected.}
    \vspace{3mm}
        \begin{tabular}{l|rr|r}
        \toprule
            Anti-pattern     & One Injection & Two Injections & Total \\
        \midrule
            AIO         & 673           & 2,487          & 3,160 \\
            EID         & 2,809         & 3,743          & 6,552 \\
            OIL         & 3             & 3              & 6     \\
            OILWI       & 3             & 0              & 3     \\
            OILWPI      & 0             & 0              & 0     \\
            UE          & 4             & 3,111          & 3,115 \\
            UEWI\_1     & 263           & 0              & 263   \\
            UEWI\_2     & 261           & 0              & 261   \\
            UEWPI       & 12            & 0              & 12    \\
            UEWIP       & 37            & 0              & 37    \\      
            SOSINETO    & 76            & 0              & 76    \\               
            OOD         & 36            & 0              & 36    \\
            OOR         & 39            & 0              & 39    \\
            CSC         & 6,344         & 0              & 6,344  \\
        \midrule
            Total       & 10,560         & 9,342           & 19,902 \\
            Total < 4,096 token & & & 10,079 \\
        \bottomrule
        \end{tabular}
        \label{tab:injection}
    \end{table}

\subsection{Translation}
\label{app:Translation}
    To leverage the language understanding capabilities of the GLM, we transform the ontologies into triples of English expressions. Examples of these transformations can be seen in \autoref{tab:translation}.
    \begin{table}[h]
        \centering
        \caption{Examples of translation rules used to transform Ontologies in the Manchester Syntax into triples in the English language.}
    \vspace{3mm}
        \begin{tabular}{ll}
        \toprule
            Manchester Syntax & Triple \\
        \midrule
            class \textsc{name} & ("\textsc{name}" "is a" "class") \\
            \textsc{ind} type \textsc{class\_name} & ("\textsc{ind}" "has class" "\textsc{class\_name}") \\
            \textsc{class1} disjoint \textsc{class2}& ("\textsc{class1}" "is disjoint with" "\textsc{class2}") \\
        \bottomrule
        \end{tabular}
        \label{tab:translation}
    \end{table}

\subsection{Prompting}
\label{sec:prompt}
    As we want to use a Llama 3 in a zero-shot setting, we have to package our task into a single prompt to give the LLM its instructions as well as the information needed to perform the task, i.¸\,e., the input. Thus, we split a single prompt into 3 parts. The prefix includes the information on the task and how the information is structured. The middle contains the information, \ie the ontology as text. The postfix defines how the output should be formed. Thus, our prompt looks as follows:
    \begin{itemize}
        \item prefix: "Determine whether the following text describes a consistent ontology. Each sentence represents a single triple.\verb|\n|\verb|\n|"
        \item postfix: "\verb|\n|\verb|\n|Answer with 0 if it is consistent and 1 if it is inconsistent.\verb|\n|Answer:"
        \item prompt: prefix + example + postfix
    \end{itemize}

\end{document}